\documentclass[letterpaper, 10 pt, conference]{ieeeconf}
\IEEEoverridecommandlockouts
\usepackage{cite}
\usepackage{amsmath,amssymb,amsfonts}
\usepackage{algorithmic}
\usepackage{graphicx}
\usepackage{textcomp}
\usepackage{xcolor}
\usepackage{svg}
\usepackage{bm}
\usepackage{url}
\usepackage{subcaption}

\title{\LARGE \bf Functionality Assessment Framework\\for Autonomous Driving Systems using Subjective Networks}

\author{Stefan Orf$^{1}$, Sven Ochs$^{1}$, Valentin Marotta$^{1}$, Oliver Conder$^{1}$, Marc René Zofka$^{1}$ and J. Marius Z\"ollner$^{1,2}$%
\thanks{
    $^{1}$FZI Research Center for Information Technology, 76131 Karlsruhe, Germany \texttt{\{orf, ochs, marotta, conder, zofka, zoellner\}@fzi.de}
}
\thanks{
    $^{2}$Applied Technical-Cognitive Systems, Karlsruhe Institute of Technology, 76131 Karlsruhe, Germany
}
}
\begin{document}

\maketitle
\thispagestyle{empty}
\pagestyle{empty}

\begin{abstract}
In complex autonomous driving (AD) software systems, the functioning of each system part is crucial for safe operation. By measuring the current functionality or operability of individual components an isolated glimpse into the system is given. Literature provides several of these detached assessments, often in the form of safety or performance measures. But dependencies, redundancies, error propagation and conflicting functionality statements do not allow for easy combination of these measures into a big picture of the functioning of the entire AD stack. Data is processed and exchanged between different components, each of which can fail, making an overall statement challenging. The lack of functionality assessment frameworks that tackle these problems underlines this complexity.

This article presents a novel framework for inferring an overall functionality statement for complex component based systems by considering their dependencies, redundancies, error propagation paths and the assessments of individual components. Our framework first incorporates a comprehensive conversion to an assessment representation of the system. The representation is based on Subjective Networks (SNs) that allow for easy identification of faulty system parts. Second, the framework offers a flexible method for computing the system's functionality while dealing with contradicting assessments about the same component and dependencies, as well as redundancies, of the system. We discuss the framework's capabilities on real-life data of our AD stack with assessments of various components.
\end{abstract}

\section{Introduction}
Performance and safety of the entire system are essential for robotics and especially AD. Measuring these metrics are infeasible due to the system's complexity. Multiple components with different tasks driven by various techniques make it hard to find a comprehensive method for such an overall assessment. Hence, it became standard practice to implement individual assessment modules (AMs) for each critical component. Such AMs are found in great numbers in literature, sometimes as part of the component or method to be monitored itself, but often as a standalone module that performs black-box monitoring.

When considering a complex system made of many different components with corresponding AMs the question arises, how the system as a whole is performing. While simple boolean logic could be applied to binary AM statements, e.g. with a logical disjunction the system's state is faulty if a single component fails, more informative statements are neglected. The question remains even when considering the functionality of AMs, which itself leads to incorrect assessments. By also taking multiple assessments about the same component into account, things get even more challenging. Assessments might support each other or be contradictory to some degree. Furthermore, in an AD system, a plethora of dependencies are present, dependencies that influence components and, of course, their assessments. In sensor data processing systems, like AD, data moves through the system from component to component, gets processed, enriched and transformed. The functionality of each downstream component is somehow dependent on components that produce the utilized data. The overall assessment becomes even more complex when considering all these challenges.

Thus, here we present a novel concept for an overall functionality assessment framework for component based message passing architectures, of course, with the complexity of AD in mind. Our framework is defined in a mathematically sound way by leveraging subjective logic (SL) and SNs \cite{josang_subjective_2016} to provide solutions to the aforementioned challenges. The framework here is adaptable, comprehensible, can deal with conflicting assessment statements and copes with dependencies, making it the ideal framework to derive a functionality statement of the entire system.

After Sec.~\ref{sec:state_of_the_art} attempts to put our work in the broader perspective of the current state of the art methods in literature, we briefly introduce SL and SNs in Sec.~\ref{sec:subjective_logic}. We use these basics to establish the foundation of our assessment framework in the form of a specialized graph, Sec.~\ref{sec:assessment_graph}. With this graph structure, we present the method for inferring an overall assessment statement in Sec.~\ref{sec:inference}, which completes our framework. Lastly, we show in Sec.~\ref{sec:evaluation} the feasibility of our framework by evaluating it on real data from our AD stack with multiple AMs.

\section{State of the Art}
\label{sec:state_of_the_art}
Substantial literature on individual AMs is available. Especially for the perception part of AD (e.g. \cite{geissler_plausibility-based_2020}, \cite{yin_ieee_2020}, \cite{hou_fault_2023}, \cite{min_fault_2023}, \cite{jin_hybrid_2024}), but also for localization (e.g. \cite{orf_modeling_2022}, \cite{seo_fail_2021}, \cite{shen_integrated_2021}) or the planning components (e.g. \cite{wang_risk_2022}, \cite{stockem_novo_self-evaluation_2023}, \cite{lu_fault_2024}). These components have in common that they are focusing only on a single part of the system. The goal of aggregating assessment of system parts is not widespread in literature.

The commonly applied industry standards ISO~26262 (functional safety in road vehicle)~\cite{functional_safety_2018} and FMEA (failure mode and effects analysis)~\cite{fmea_2018} aim at the identification and mitigation of risks in automotive systems to increase safety. But while ISO~26262 defines a safety life cycle and FMEA is a general method for systematically determining potential failures. They don't provide a technical framework to incorporate individual safety assessments of software components while at the same time providing aggregation strategies and reasoning about the whole system on a technical level during the vehicle's runtime. Such assessment frameworks, especially in AD, are sparse in the literature. 

However, the modular fault diagnosis framework proposed in \cite{orf_modular_2024} outlines the general concept of assessment aggregation, together with a classification scheme for faults in component based AD architectures, but this concept is limited due to its rule based approach. Another approach is \cite{youssef_general_2018} that leverages simple logic circuits for its diagnosis system. 
We argue that our approach is more powerful as it allows AMs to utilize different data sources, and therefore have complex dependencies, instead of only being dependent on the component under test. Furthermore, our framework provides the possibility to group AMs with additional assessments, thereby allowing a clear hierarchical assessment structure. A precise adjustment for enhancing or downgrading AMs is also given. In contrast, we show the feasibility of our approach on real world data from our AD vehicle. Beyond that, we see comprehensibility and clarity through the bijective mapping of our assessment structure to the monitored system as the key advantage of our framework.

\section{Subjective Logic}
\label{sec:subjective_logic}
SL is an extension to probabilistic logic and was invented by Audun J\o{}sang \cite{josang_subjective_2016}. The main advantage of SL over traditional probabilistic methods is the possibility of modelling the presence/absence of knowledge in the form of an explicit uncertainty specification. A brief overview of Subjective Logic (SL) is provided herein; for a comprehensive treatment, the reader is referred to \cite{josang_subjective_2016}. \cite{josang_subjective_2016} thoroughly covers SL basics, as well as many interesting topics, like a plethora of operators, subjective trust and SNs. We also follow the notation of \cite{josang_subjective_2016} where possible,  to facilitate continuity and comparison with the original work. We also want to highlight our own SL library, named \emph{SUBJ} \cite{orf_subj_2025}, which is written in C++ and comes with Python bindings. Our assessment framework implementation presented in this article is built on SUBJ. Another helpful resource is \cite{josang_subjective_2020}, which provides web-based visualizations and operator demonstrations.

\subsection{Opinions}
The basic elements of SL are called subjective \emph{opinions}. First of all, an opinion~$\omega_x^A$ of an \emph{owner} $A$ defines \emph{belief} mass $\bm{b}_x$ over a given state space or \emph{domain} $\mathbb{X}$ with cardinality $k = |\mathbb{X}|$ in respect of a random variable $x$. The elements of $\mathbb{X}$ are assumed to be exclusive and exhaustive. The belief mass $\bm{b}_x : \mathbb{X} \rightarrow [0,1]$ assigned to each state of $\mathbb{X}$ corresponds to classic probability and expresses support in the random variable $x$ being in this state. To reflect subjectivity, an opinion $\omega_x^A$ defines an \emph{uncertainty} $u_x \in [0,1]$, that expresses how certain the holder or owner $A$ of this opinion is about this statement. There is the additivity requirement $u_x + \sum_{i \in \mathbb{X}} \bm{b}_x(i) = 1$. An opinion also has a \emph{base rate} $\bm{a}_x: \mathbb{X} \rightarrow [0,1]$, which can be seen as the prior probability of $x$. Again, there is an additivity requirement $\sum_{i \in \mathbb{X}} \bm{a}_x(i) = 1$. An opinion $\omega_x^A$ of an entity $A$ over a random variable $x$ is then defined as
\begin{equation}
    \omega_x^A = (\bm{b}_x, u_x, \bm{a}_x).
\end{equation}

There exists a bijective mapping between a Dirichlet probability density function (PDF) and an opinion. Furthermore, Dirichlet PDFs can be represented in an evidence notation, where a vector of evidence for each state determines the PDF. This comes in handy when generating opinions from evidence. Opinions with high uncertainty are mapped to a Dirichlet PDF with higher variance,
while those with low uncertainty have a narrower shape
.
Opinions over a domain $\mathbb{X}$ with cardinality $k = 2$ are called \emph{binomial} opinions. Following \cite{josang_subjective_2016}, the belief $b_x$ and disbelief $d_x$ can be explicitly stated when notating a binomial opinion $\omega_x^A = (b_x, d_x, u_x, a_x)$, where $a_x$ denotes only the first element of the base-rate, due to the additivity requirement. In the assessment framework, most opinions are assumed to be binomial. These opinions map to a Beta PDF, which is a 2-dimensional Dirichlet PDF.

\subsection{Trust}
\label{sec:trust}
The concept of \emph{trust} of an entity $A$ in another entity $B$ represents the degree of confidence $A$ puts in the statement (or opinion) of $B$ about the functionality of $x\in \mathbb{X}$. Thus, $A$ can derive an own opinion about $x$ via its trust in $B$. Trust can also be modeled with SL by utilizing binomial opinions. The opinion $\omega_x^B$ of $B$ about the outcome of $x$ is then called \emph{functional trust}. Whereas $\omega_B^A$ is the \emph{referral trust} of $A$ in $B$. By using the opinion of $B$ and the \emph{trust discount} operator~$\otimes$ $A$ can then derive itself functional trust in $x$ with ${\omega_x^A = \omega_B^A \otimes \omega_x^B}$ \cite{josang_subjective_2016}. With this operator the resulting $\omega_x^A$ responds more to $\omega_x^B$ if $A$ trusts $B$ with high belief and low uncertainty. In the assessment framework, trust is used to handle assessments differently, e.g., by emphasizing important assessments.

\subsection{Belief Fusion}
\label{sec:belief_fusion}
Each opinion is held by an owner, making it possible to have opinions about the same subject by different entities. In the assessment framework assessment opinions about the same functional component from different entities may exist. Thus, a method for fusion is needed. SL provides several fusion operators, from which we utilize the \emph{aleatory cumulative belief fusion} operator~$\oplus$ \cite{josang_subjective_2016}. A resulting opinion ${\omega_x^{A\diamond B} = \omega_x^A \oplus \omega_x^B}$ corresponds more to the opinion $\omega_x^A$ or $\omega_x^B$ with lower uncertainty.

\subsection{Subjective Networks}
\label{sec:subjective_networks}
In \cite{josang_subjective_2016}, the concept of subjective networks is described as a combination of trust networks with Bayes nets, where one or both parts are assumed to be subjective. Subjective trust networks model entities with their trust relationships as opinions. In subjective Bayes nets conditional probabilities between random variables are modeled with the conditionals also being subjective opinions. It is then possible for an entity $A$ to draw conclusions, e.g. to employ predictive reasoning by incorporating the functional and referral trust of entities that have an opinion about these variables. In our assessment framework, we leverage subjective trust networks for coping with assessments from multiple, potentially conflicting sources and subjective Bayes nets for dependencies between functional components. In these parts, trust and probabilities are modeled as opinions.

Deriving trust is explained in Sec.~\ref{sec:trust}. Performing predictive reasoning of the state of variables with evidence on other variables and assumptions on conditionals in the subjective Bayes net can be carried out with the deduction operator $\circledcirc$~\cite{josang_subjective_2016}. For a joint opinion of dependency variables $\omega_X$ and a set of opinions $\bm{\omega}_{y | X}$, which describes how each combination of parent states affects the states of $y$, $\omega_{y || X} = \omega_X \circledcirc \bm{\omega}_{y | X}$ is then the deduced opinion.

\section{Assessment Graph}
\label{sec:assessment_graph}
The assessment framework leverages SL to infer a statement of functionality of the entire system. The inference (see Sec.~\ref{sec:inference}) is done based on the \emph{assessment graph} (AG) that is composed of nodes for all functional components of the system, as well as nodes for the AM. In this section we describe the construction of the AG for component based message passing systems (which is a common architecture in e.g. AD or robotics). We begin by stating the preliminaries of the system to be assessed, with its components and dependencies, then we elaborate on the creation of the two-fold AG, which is based on SNs.

The AG is a SN that is comprised of nodes and different types of edges between them. The advantage of this representation is its descriptiveness and comprehensibility by a bijective mapping of the system's component based architecture to the AG. The AG $S = (V, E)$ is defined as a set of vertices or nodes $V$ and directed edges $E \subseteq V \times V$. To distinguish between functional nodes $V_f$ and AMs $V_a$ the set of vertices is $V = V_f \cup V_a$. The same holds for edges $E = E_d \cup E_{ft} \cup E_{rt}$ with $E_d$, $E_{ft}$ and $E_{rt}$ being edges that describe dependencies, functional trust and referral trust respectively. The AG is then
\begin{equation}
\label{eq:assessment_graph}
    S = (V_f \cup V_a\ ,\  E_d \cup E_{ft} \cup E_{rt}).
\end{equation}
The advantage of the AG lies in its direct correspondence to the system's architecture. From a given AG the relevant system parts, i.e. fault locations, can easily be determined. The mapping from a component based system architecture with AMs to an AG of Eq.~\ref{eq:assessment_graph} is described in the following.

\begin{figure*}[tbp]
    \centering
    \vspace{2mm}
    \includegraphics[width=1.0\textwidth]{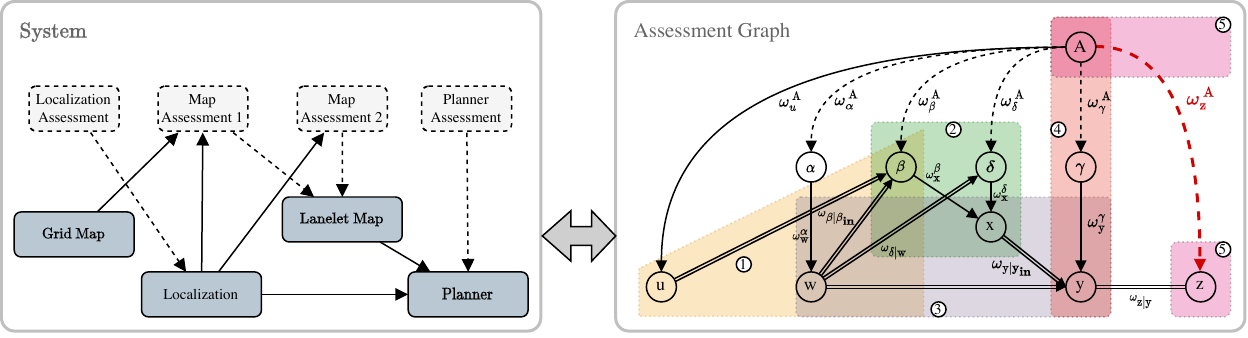}
    \caption{Example system of multiple functional components depicted on the left with dependencies by message passing (solid lines) together with assessment modules each monitoring a component (dashed lines). AG with nodes representing the components on the left. Dependencies (double lines), functional trust of AMs (solid lines), and referal trust of the overall assessment $A$ in the AMs (dashed lines) are shown. Key aspects of the assessment framework are given by ability to model dependencies of AMs \raisebox{.5pt}{\textcircled{\raisebox{-.9pt} {1}}} (see Sec.~\ref{sec:opinions_on_dependencies}), concurrent AMs \raisebox{.5pt}{\textcircled{\raisebox{-.9pt} {2}}} (see Sec.~\ref{sec:concurrent_assessments}), general dependencies \raisebox{.5pt}{\textcircled{\raisebox{-.9pt} {3}}} (also Sec.~\ref{sec:opinions_on_dependencies}), derivation of trust \raisebox{.5pt}{\textcircled{\raisebox{-.9pt} {4}}} (see Sec.~\ref{sec:trusting_assessments}) and inferring an overall assessment opinion $\omega_Z^A$ (see Sec.~\ref{sec:algorithm}).}
    \label{fig:assessment_graph}
\end{figure*}

\subsection{Functional Components}
Complex systems that pass messages between components imply dependencies that are best described with a dependency graph (see Fig.~\ref{fig:assessment_graph}). In AD, or robotics in general, usually sensor data is collected at one end of the system and then passed through and enriched by multiple other components until an output decision (e.g. controls to drive the vehicle) is generated. This leads us to the assumption that dependencies are fully described by the data exchange, meaning that the functionality of a component (e.g. Planner in Fig.~\ref{fig:assessment_graph}) directly depends, besides it's own operability, only on the functioning of parents (e.g. Lanelet Map and Localization in Fig.~\ref{fig:assessment_graph}) that generate the data the component is processing. We note here that it is possible to use other types of dependency in the framework, which then needs to be reflected in the AG likewise.

A component of the system is characterized by its functionality of processing input data and producing output data. Components can also have no input, e.g. sensors are typical components without any input connection, which use the environment outside the system for their calculations. Components without any outgoing connections are also possible. Thus, a component is defined as an entity $x^*$ that processes data from a (possibly empty) set of components $\{p_1^*,\dots\}$ and outputs it's data to a (also possibly empty) set of other components~$\{c_1^*,\dots\}$. 

For each component $x^*$ the AG $S$ contains a node $x \in V_d$. 
The set $\bm{x}_{\text{out}} = \{ c \in V_d | (x,d)\in E_d \}$ represents all directly dependent nodes (or direct children) of $x$ where their respective components process the data of $x$.
This implies that directed edges from nodes, whose corresponding components produce the data $x^*$ is processing, i.e. $x$'s direct parents, are also included. This set of nodes $x$ directly depends on is then $\bm{x}_{\text{in}} = \{ d \in V_d | (d,x) \in E_d\}$.
It is obvious that for any node $x$ with $y \in \bm{x}_{\text{out}} \Leftrightarrow x \in \bm{y}_{\text{in}}$ and vice versa for it's parent nodes. We also define the set of all (direct and indirect) parent nodes recursive as $\bm{\hat{x}}_{\text{in}} = \{y \in \bm{\hat{p}}_{\text{in}} , \forall p \in \bm{x}_{\text{in}} \}$. Likewise, the set of all (direct and indirect) children nodes of $x$ is $\bm{\hat{x}}_{\text{out}} = \{y \in \bm{\hat{c}}_{\text{out}} , \forall c \in \bm{x}_{\text{out}} \}$. We rely on a loop free dependency graph, meaning that $x$ can not depend on any node that corresponds to components which indirectly processes data from $x^*$ and thus, $\bm{\hat{x}}_{\text{in}} \cap \bm{\hat{x}}_{\text{out}} \overset{!}{=} \emptyset$.

\subsection{Assessment Components}
Similar to the functional components in the system that process data, there are AMs that monitor the functioning of functional components or system parts, but do not contribute to the functionality of the system. The framework relies on these AMs to infer an overall functionality statement of the system. The AG $S$ also contains nodes $\alpha \in V_a$ for each AM $\alpha^*$. An AM $\alpha^*$ is characterized by not outputting data to the system, which means $\alpha \notin \bm{\hat{x}}_{\text{in}}, \forall x \in V_f$ and $E_d \neq V_a \times V_f$. But rather, each AM $\alpha^*$ monitors a functional component $x^*$. This connection is modeled by using functional trust of $\alpha^*$ in $x^*$. This is reflected by an edge $(\alpha, x) \in E_{\text{ft}}$, with the set of functional trust edged being $E_{\text{ft}} = V_a \times V_f$. Note that there might be multiple AMs that monitor a distinct node. We define $\bm{\textbf{As}}_x$ as the set of all AMs that monitor $x$.

Most AMs generate an assessment about a functional component by utilizing data from other parts of the system. The reliability of an AM is thus dependent on the functioning of the components whose data it uses. Thus, for each $\alpha \in V_a$, $E_d$ is extended by $(x, \alpha) \in V_f \times V_a ,\forall x$ whose data $\alpha$ is utilizing. To keep a loop free dependency graph, we do not include dependencies of an AM to its assessed node, though, it most likely uses its data and, thus, is dependent on it. 

There might also be an AM $\beta^*$ that measures the functionality of an AM $\alpha^*$. In this case an edge $(\beta, \alpha) \in E_{\text{ft}}$ with both $\beta, \alpha \in V_a$ exists. While this case is possible, we argue that it is rare and don't go into detail for brevity reasons. But the calculations below can easily be extended for this case.

\subsection{System Assessment Nodes}
Some AMs can be more trusted than others. This can be reflected by allowing different referral trusts in all AMs. For this, an artificial node $A$ has to be included in $V_a$. We call $A$ the \emph{overall assessment}. This node can be seen as the assessment entity for the complete system or the system part under test. Referral trust is represented in an AG $S$ by edges $(A, \alpha) \in E_{\text{rt}}, \forall \alpha \in V_a$. Also, for functional nodes that don't have at least one AM that generates a functionality statement about it, the overall assessment needs to provide this functionality statement. There are different ways such a statement could be designed. E.g. there might be expert assumptions like a vacuous or dogmatic opinion which states absolute un- or certainty that the component is functioning or failing. Of course, varying uncertainties are also possible. To utilize such expert opinions functional trust edges $(A, x) \in E_{\text{ft}} : \neg\exists (\alpha, x) \in E_{\text{ft}}, \forall \alpha \in V_a$ have to be included in $S$. Additionally, there might be additional functional trust from $A$ to any $x\in V_f$, with which the existing assessment opinions about $x$ can be extended.

The AG is completed by an additional artificial node $Z \in V_f$. This artificial node is created such that it is the ultimate node that is dependent on all other functional nodes. When assuming dependencies through message connections, $Z$ can be seen as a sink for all output messages of the system or the system part under test. With the assumption that the system's functionality is reflected by the functioning of its components that don't output data (i.e. they output data to the environment), the functionality statement of $Z$ represents the system's functionality. The corresponding edges $(x, Z) \in E_d, \forall x\in V_f: \bm{x}_{\text{out}} = \emptyset$ have then also to be added to $S$.

\subsection{Assessments of Exchanged Data}
It is also possible to represent assessments that measure the correctness (or "functionality") of the exchanged data. Again, the assumption is that the receiving component's functionality depends on the correctness of the data, which in turn depends on the functioning of the sending component. The necessary change to the functionality AG is then straightforward and involves inserting an additional node $a$ for the data exchanged between an sender $x^*$ and all receivers of this data. Edges for the sending $(x, a) \in E_d$ and receiving nodes $(a, y) \in E_d, \forall y \in \bm{x}_{\text{out}}$ need to be inserted accordingly.

\section{Overall Assessment}
\label{sec:inference}

The ultimate goal of the functionality assessment framework is to generate a statement on the functionality of the assessed system (or system part). Such a statement can be created by leveraging SL. As mentioned above, the statement for the overall functionality of the system corresponds to the assessment of the artificial node $Z \in V_f$. The AG $S$ is two-fold with the dependencies described by $(V, E_d)$ and the assessment part described by $(V, E_{\text{ft}} \cup E_{\text{rt}})$. Note, that we model dependencies of the AMs to the functional components, which are also included in $E_d$. This bipartite graph corresponds to an SN, with the dependencies being the subjective Bayes net and the assessment part being the subjective trust network. By considering $S$ as an SN, we can use subjective opinions and SL operators to arrive at an overall assessment opinion. This procedure is explained in the following.

\subsection{Trusting Assessments}
\label{sec:trusting_assessments}
Each AM node $\alpha \in V_a$ is assumed to have a functionality opinion $\omega_x^\alpha$ about a component node $x\in V_f$. The artificial node $A$ potentially has multiple such opinions for all missing assessments. The opinion $\omega_x^\alpha = (b_x, d_x, u_x, a_x)$ is modeled as a binomial opinion with the domain ${\mathbb{X} = \{\text{\emph{functional}}, \text{\emph{non-functional}}\}}$ and reflects the functionality opinion of $\alpha$ about $x$. In $S$ each edge $(\alpha, x) \in E_{\text{ft}}$ corresponds to such a binomial opinion $\omega_x^\alpha$.

The assessment framework features trusting individual AMs differently. This is achieved by defining referral trust opinions $\omega_\alpha^A = (b_\alpha, d_\alpha, u_\alpha, a_\alpha)$ of the overall assessment $A$ to all AMs $\alpha\in V_a$. The domain for referral trust is also $\mathbb{X}$, where the interpretation of the two states are assumed to be that the entity trusts or distrusts an AM in its functionality assessment. For each edge $(\alpha, \beta) \in E_{\text{rt}}$ an opinion $\omega_\beta^\alpha$ is assumed.

It is possible to have multi-level referral trust, i.e. a node $\alpha$ that has a trust opinion $\omega_\beta^\alpha$ in a component $\beta$ that also has trust $\omega_x^\beta$ in another component $x$, etc. In this work, we avoid multi-level trust and only use first-level trust $\omega_\alpha^A, \forall \alpha \in V_a \setminus \{A\}$ of the overall assessment node $A$ to all assessment nodes without itself. Nevertheless, we don't exclude multi-level trust from the framework, e.g. for functionality assessments of multiple nested parts of the system. The following calculations also hold for this case.

For $A$ to have a functional trust in a component $x \in V_f$ we employ the trust discount operator, as in Sec.~\ref{sec:trust}. Assume an assessment node $\alpha \in V_a \setminus \{A\}$, with $\alpha \notin \bm{y}_{\text{out}}, \forall y\in V_f$. Further assume that $\alpha$ has functional trust $\omega_x^\alpha$ in $x \in V_f$ and the referral trust $\omega_\alpha^A$ of $A$ in $\alpha$, then
\begin{equation}
\label{eq:trust_discount}
    \omega_x^{[A;\alpha]} = \omega_\alpha^A \otimes \omega_x^\alpha
\end{equation}
describes the derived functional trust of $A$ in $x$ via $\alpha$. With this referral trust $\omega_\alpha^A$ the AM $\alpha^*$ can be degraded or enhanced. In Fig.~\ref{fig:algorithm} functional trust can be derived for nodes $\alpha_0$ and $\alpha_1$.
Since, for each pair $(\alpha, x) \in \{V_a \setminus A\} \times V_f$ at most one $(\alpha, x) \in E_{\text{ft}}$ exists, the application of the trust discount operator yields an $\omega_x^{[A;\alpha]}, \forall (\alpha,x) \in E_{\text{ft}}$ with $\alpha \neq A$. Note, that assessment nodes that depend on functional nodes are described later in this section.

\subsection{Concurrent Assessments}
\label{sec:concurrent_assessments}
There can be concurrent assessments either from different AMs with nodes $\alpha, \beta \in V_a$, or from the overall assessment $A$ for a single functional component $x \in V_f$. All these assessments are represented as a binomial opinion and describe the functionality of $x$. These opinions can be highly contradictory, which is why aleatory cumulative belief fusion is applied in the assessment framework, as described in Sec.~\ref{sec:belief_fusion}. For two different functional trust opinions $\omega_x^\alpha, \omega_x^\beta$ of $\alpha, \beta \in V_a$ for $x\in V_f$ the fusion is then defined as
\begin{equation}
\label{eq:trust_fusion}
    \omega_x^{(\alpha \diamond \beta)} = \omega_x^\alpha \oplus \omega_x^\beta.
\end{equation}
This also holds if the functional trust is derived. E.g. for $A$ that has multiple derived functional trusts in $x$ via $\alpha$ and $\beta$. The fused functional trust is then $\omega_x^{([A;\alpha] \diamond [A;\beta])}$. Note that the aleatory cumulative fusion operator is associative and, thus, can be applied multiple times if there are more than two opinions.

\begin{figure}[htbp]
    \vspace{2mm}
    \centering
    \includegraphics[width=0.58\linewidth]{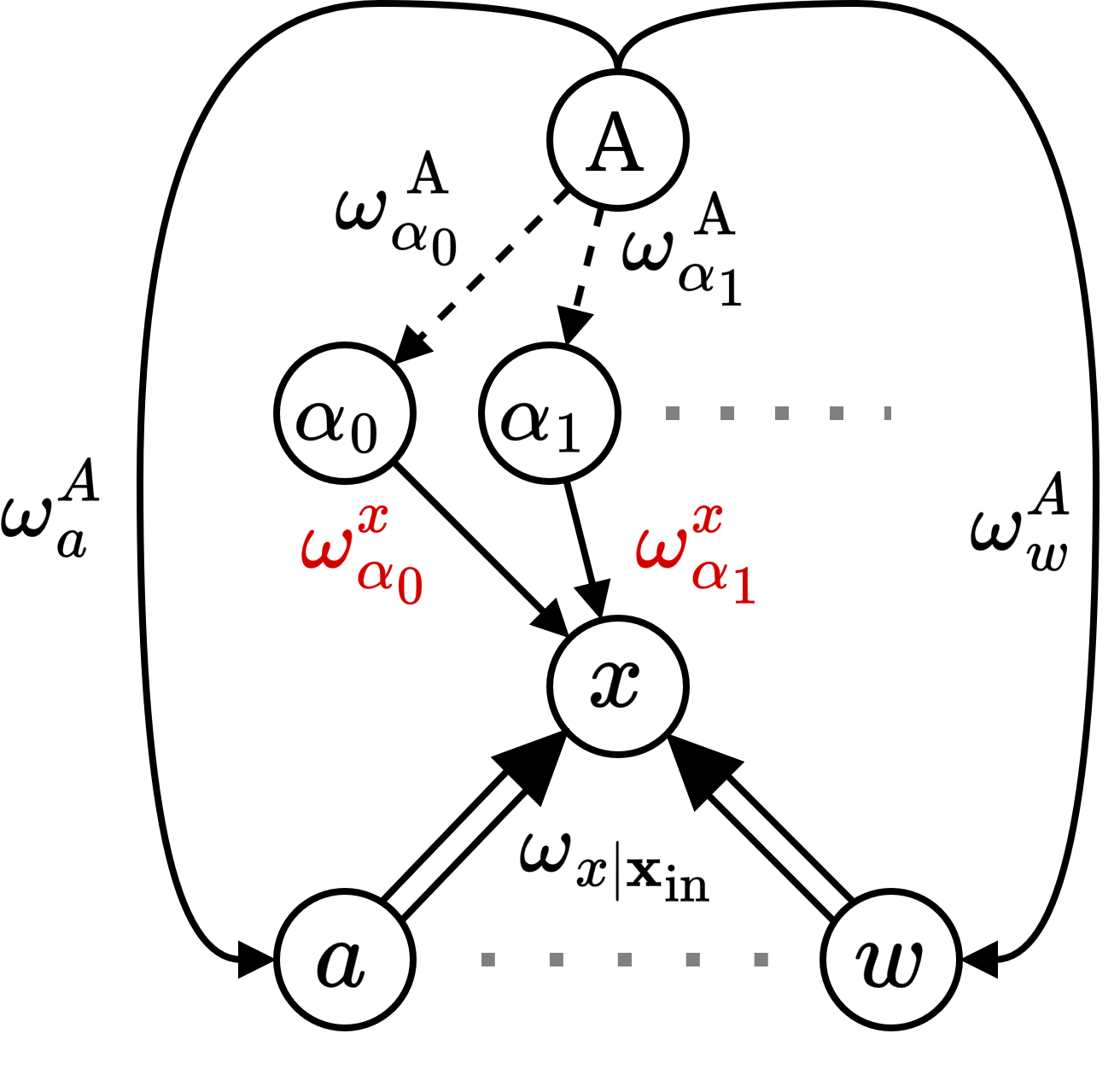}
    \caption{Depiction of the recursive Eq.~\ref{eq:general_assessment} for calculation of an opinion $\omega_x^A$ representing operability of node $x$ of the overall assessment $A$ as a graph. AMs $\alpha_0, \alpha_1,\dots$ each having an opinion $\omega_x^{\alpha_0}, \omega_x^{\alpha_1}, \dots$ on $x$, with an opinion $\omega_{\alpha_0}^A, \omega_{\alpha_1}^A,\dots$ from $A$ that represents the trust in the AMs. Opinions $\omega_a^A,\dots,\omega_w^A$ on the dependencies $a,\dots,w \in \bm{x}_{\text{in}}$ reflect their operability.}
    \label{fig:algorithm}
\end{figure}

The typical case for concurrent assessments is depicted in Fig.~\ref{fig:algorithm}, where two assessment nodes $\alpha_0, \alpha_1$ have functional trust in a single component $x$, while $A$ has referral trust in both assessments. By first applying the trust discount operator (see Sec.~\ref{sec:trusting_assessments}) two referral trust opinions $\omega_x^{[A;\alpha_0]}, \omega_x^{[A;\alpha_1]}$ in $x$ are calculated, which are then fused together with the fusion in Eq.~\ref{eq:trust_fusion} to form $\omega_x^A$.

\subsection{Opinions on Dependencies}
\label{sec:opinions_on_dependencies}
In the assessment framework, it is assumed that the functionality of a component influences the functionality of its dependent components. With the subjective Bayes net of the AG the overall assessment node $A$ can infer a functional trust opinion $\omega_{x || \bm{x}_{\text{in}}}^A$ about a functional node $x\in V_f$ given all its dependencies $\bm{x}_{\text{in}}$ with the deduction operator (see Sec.~\ref{sec:subjective_networks}). For this, $A$ needs to hold a set of opinions $\bm{\omega}_{x | \bm{x}_{\text{in}}}^A$ about how each combination of the functioning of dependency nodes, expressed as a joint opinion $\omega_{\bm{x}_{\text{in}}}^A$, influences $x$. In case all dependencies of $x$ are independent of each other the joint opinion $\omega_{\bm{x}_{\text{in}}}^A$ can be calculated with multinomial multiplication (see Sec.~\ref{sec:subjective_networks}), which means
\begin{equation}
\label{eq:dependencies_multiplication}
    \omega_{\bm{x}_{\text{in}}}^A = \prod_{y \in \bm{x}_{\text{in}}} \omega_y^A. 
\end{equation}
For a node $x\in V_f$ the deduced opinion $\omega_{x || \bm{x}_{\text{in}}}^A$ of $A$ given $x$'s dependencies $\bm{x}_{\text{in}}$, together with a set of opinions $\bm{\omega}_{x | \bm{x}_{\text{in}}}^A$ about how the functioning of the dependencies influences the functioning of $x$, is calculated with the deduction operator
\begin{equation}
\label{eq:deduction}
    \omega_{x || \bm{x}_{\text{in}}}^A = \omega_{\bm{x}_{\text{in}}}^A \circledcirc \bm{\omega}_{x | \bm{x}_{\text{in}}}^A.
\end{equation}
The resulting opinion $\omega_{x || \bm{x}_{\text{in}}}^A$ is then also a binomial opinion (see also Fig.~\ref{fig:algorithm}). Likewise the deduction operator is used in the same way to generate $\omega_{\alpha || \bm{\alpha}_{\text{in}}}^A$ for AMs $\alpha \in V_a$.

\subsection{Algorithm}
\label{sec:algorithm}
In general, to arrive at an assessment opinion on any node with its dependencies and assessments a combination of the operators above has to be performed. Given a node $x\in V_f$, its dependencies $\bm{x}_{\text{in}}$, an opinion $\omega_{\bm{x}_{\text{in}}}^A$ about the functionality of these dependencies, a set of opinions $\bm{\omega}_{x | \bm{x}_{\text{in}}}^A$ on how the functionality of the dependencies influences $x$, assessment nodes $\bm{\textbf{As}}_x$ of $x$, the assessment opinions $\omega_x^\alpha, \forall \alpha \in \bm{\textbf{As}}_x$ and respective trust opinions $\omega_\alpha^A$, then, by applying Eqs.~\ref{eq:trust_discount},~\ref{eq:trust_fusion},~\ref{eq:deduction} the assessment opinion about $x$ is given by
\begin{equation}
\begin{aligned}
\label{eq:general_assessment}
    \omega_x^A &= \omega_{x || \bm{x}_{\text{in}}}^A \oplus \omega_x^{([A; \alpha_0]\diamond \dots)} \\
               &= \omega_{x || \bm{x}_{\text{in}}}^A \oplus (\omega_x^{[A; \alpha_0]} \oplus \dots) \\
               &= \omega_{x || \bm{x}_{\text{in}}}^A \oplus ((\omega_{\alpha_0}^A \otimes \omega_x^{\alpha_0}) \oplus \dots) \\
               &= (\omega_{\bm{x}_{\text{in}}}^A \circledcirc \bm{\omega}_{x | \bm{x}_{\text{in}}}^A ) \oplus (\bigoplus_{\alpha \in \bm{\textbf{As}}_x} \omega_\alpha^A \otimes \omega_x^\alpha).
\end{aligned}
\end{equation}
If the dependencies of $x$ are independent from each other Eq.~\ref{eq:dependencies_multiplication} can be used to calculate $\omega_{\bm{x}_{\text{in}}}^A$. The procedure is also visualized in Fig.~\ref{fig:algorithm}.

Getting an overall assessment about the system is then simply achieved by generating the opinion $\omega_Z^A$ of $A$ about the artificial node $Z$ that is dependent from all other nodes in $V_f$. With the given set of opinions $\bm{\omega}_{Z || \bm{Z}_{\text{in}}}^A$ on how the overall assessment is influenced by the end-point nodes, the overall assessment can be calculated with
\begin{equation}
    \omega_Z^A = \omega_{\bm{Z}_{\text{in}}}^A \circledcirc \bm{\omega}_{Z | \bm{Z}_{\text{in}}}^A.
\end{equation}
Again, by assuming independence the dependency opinion is given by Eq.~\ref{eq:dependencies_multiplication}. The assessment opinions $\omega_x^A, \forall x\in \bm{Z}_{\text{in}}$, in turn, are given by recursively applying Eq.~\ref{eq:general_assessment}.

\subsection{Missing Assessments/Opinions}
The overall assessment relies on individual AMs that produce subjective opinions on all nodes $x\in V_f$. But there might be cases where creating these opinions is not feasible or these opinions might be missing otherwise. To cope with this, $A$ can assume a constant assessment opinion $\omega_x^A = \text{const.}$ about a node $x\in V_f$ that has no assessment $\alpha\in V_a$. Such a constant opinion can be, e.g., a vacuous opinion with maximum uncertainty. But of course different values for the uncertainty and the assumed functionality of $x$ are possible. 

If there are AMs that produce a non-SL assessment, it can not easily be used in the assessment framework. In this case a mapping of the original domain to binomial opinions is required. Functionality statements can easily, possibly by using thresholds, be translated into the binomial domain. The difficulty lies within determining a proper uncertainty. Again, a constant uncertainty can be assumed. However, with the mapping to a Beta PDF in evidence notation SL provides the right tool for this purpose. One would then combine several assessment measurements to create an opinion, whose uncertainty is determined by the number of measurements.

But of course, the overall assessment profits from a high coverage of the functional components with AMs. But it is possible to cope with minor absences of adequate functionality assessments.

\begin{figure*}[htbp]
    \vspace{2mm}
    \begin{subfigure}[t]{0.33\textwidth}
        \centering
        \includegraphics[width=1.0\linewidth]{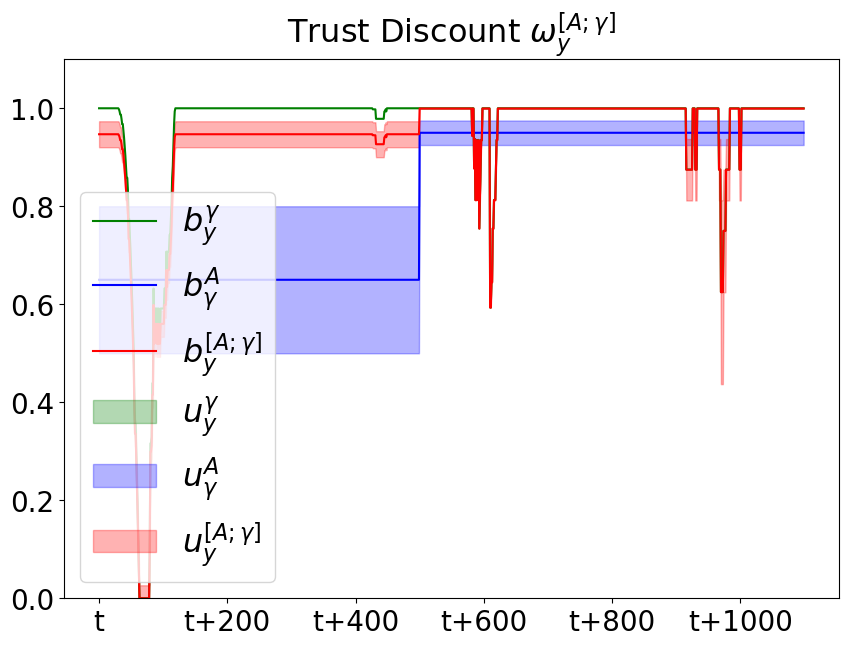}
        \caption{}
        \label{fig:trust_discount_plot}
    \end{subfigure}
    \begin{subfigure}[t]{0.33\textwidth}
        \centering
        \includegraphics[width=1.0\linewidth]{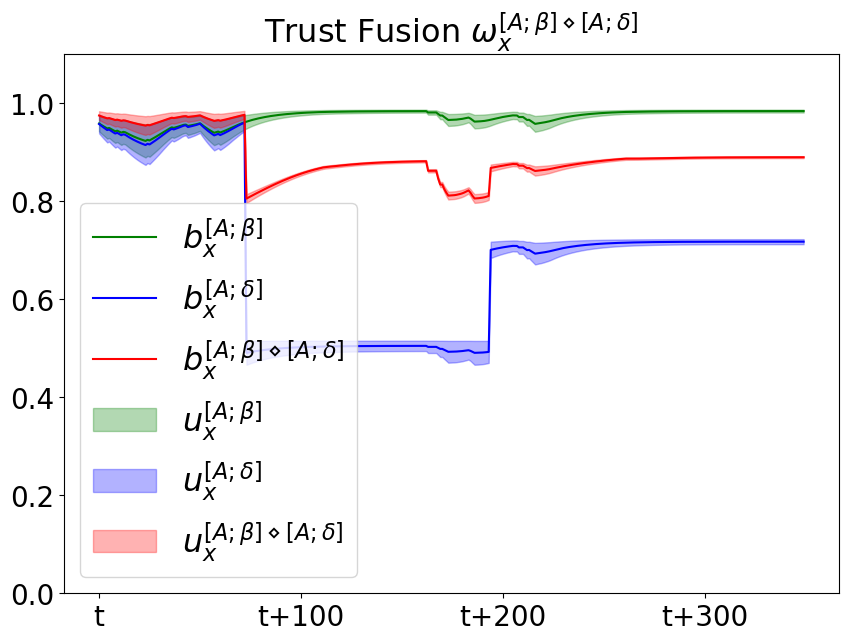}
        \caption{}
        \label{fig:trust_fusion_plot}
    \end{subfigure}
    \begin{subfigure}[t]{0.33\textwidth}
        \centering
        \includegraphics[width=1.0\linewidth]{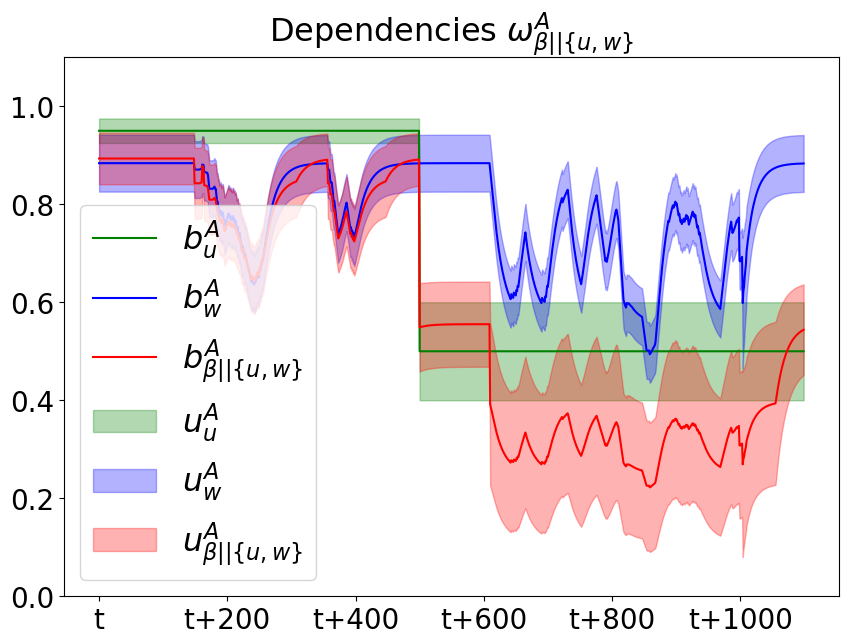}
        \caption{}
        \label{fig:dependencies_plot}
    \end{subfigure}
    \caption{Changes of beliefs $b$ (lines) over time with corresponding uncertainty $u$ (height of the shaded area) during the inference of $\omega_Z^A$ for opinions marked as \raisebox{.5pt}{\textcircled{\raisebox{-.9pt} {4}}}, \raisebox{.5pt}{\textcircled{\raisebox{-.9pt} {2}}}, \raisebox{.5pt}{\textcircled{\raisebox{-.9pt} {1}}} in the AG in Fig.~\ref{fig:assessment_graph}. The resulting trust discounted belief (red) in (a) is significantly lowered, compared to the input opinion's belief (green), if the trust (blue) is low. In (b) it is clearly visible that both input opinion's beliefs (green, blue) influence the resulting opinion's belief (red). Deduction (c) of a resulting opinion given two opinions (green, blue), where functionality of both inputs is required.}
    \label{}
\end{figure*}

\section{Evaluation}
\label{sec:evaluation}
The AG $S$ in Fig.~\ref{fig:assessment_graph} is used to evaluate the feasability of an overall assessment with the presented assessment framework. Several AMs are included in $S$. A grid map $u^*$, without a distinct assessment, a localization system $w^*$ with AM $\alpha^*$, the Lanelet map \cite{poggenhans_lanelet2_2018} with two AMs $\beta^*$ and $\delta^*$, and the planner module $y^*$ with AM $\gamma^*$ form the system that is evaluated. In the following we briefly describe the AMs, after which we discuss the effects of trust discount for AMs, trust fusion for concurrent AMs, deduction for dependencies, as well as the overall assessment over time. The evaluation is done on real data of our AD vehicle CoCarNextGen \cite{heinrich_cocarnextgen_2024}.

\subsection{Self Assessments}

\subsubsection{Trajectory Planner Assessment} \label{sec:planner_assessment}
One core AM in the AD stack used is the Trajectory Planner AM $\gamma^*$, monitoring the Trajectory Planner $y^*$. This is represented in the AG $S$ by a node $\gamma\in V_a$ and an edge $(\gamma, y) \in E_{\text{ft}}$. The functional trust $\omega_y^\gamma$  corresponding to the edge $(\gamma, y) \in E_{\text{ft}}$ is used to monitor $y^*$ and detect a potential hazardous trajectory $\bm{T} = \{p_0,...,p_{n-1}\}$ as the output of $y^*$. The best-fit lanelet sequence on a lanelet map \cite{poggenhans_lanelet2_2018} is calculated. Within this lanelet sequence, the distance to the centerline of the lanelet is calculated for each trajectory point $p_i \in \bm{T}$ and compared with a previously learned distribution of distances for that point. Since the evaluation depends on the sequence of detected lanelets, the uncertainty $u_x \in [0,1]$ increases with each combination of potential lanelets. Every trajectory point is weighted with $w_i = \frac{n-i}{n}$ and assigned to sets $\bm{F}$ or $\bm{N}$, depending whether it is $\text{\emph{functional}}$ or $\text{\emph{nonfunctional}}$ according to the learned distributions respectively, with $\bm{F}\cup\bm{N} =\bm{T}$ and $\bm{F}\cap\bm{N} =\emptyset$. The functional trust $\omega_y^\gamma=(b_x,d_x,u_x)$ is calculated with
\begin{align}
    b_x &= (1 - u_x) \cdot (\sum\limits_{p_i \in \bm{F}} w_i) \cdot (\sum\limits_{j=0}^{n-1} w_j)^{-1},\\
    d_x &= (1 - u_x) \cdot (\sum\limits_{p_i \in \bm{N}} w_i) \cdot (\sum\limits_{j=0}^{n-1} w_j)^{-1}.
\end{align}

\subsubsection{Lanelet Map Assessments}
Two distinct AMs $\beta^*$ and $\delta^*$ are employed to assess the local quality of the Lanelets~\cite{poggenhans_lanelet2_2018} $x^*$ at the vehicle's position which is provided by the localization $w^*$. The AM $\delta^*$ assesses the Lanelet Map isolated without any additional data as displayed in the AG $S$ by the edge $(\delta,x) \in E_{ft}$ (See Fig.~\ref{eq:assessment_graph}). This facilitates both the detection of definitive errors using predefined conditions and the evaluation of the map's plausibility based on the characteristics of the local Lanelets such as the variance in width. The belief of the functional trust $\omega_x^{\delta}$ corresponding to the edge $(\delta,x) \in E_{ft}$ describes the estimated probability of the Lanelet Map being locally valid.
The AM $\beta^*$ employs data from an occupancy grid map $u^*$ to compare the Lanelet Map with sensory data and to evaluate the correctness of the placement of the Lanelets, as displayed by the edge $(\beta,x) \in E_{ft}$ in AG $S$. The belief of the functional trust $\omega_x^{\beta}$ describes the estimated probability of the local Lanelets being placed correctly based on the grid map $u^*$.
In both AMs, opinions are formed on the basis of evidence derived from the previous track record of the methods used. The underlying evidence reflects the extent to which the methods applied have classified Lanelets correctly or incorrectly in the past.
Both AMs $\beta$ and $\delta$ use the localization module and are therefore also dependent on $w$.

\subsubsection{Localization Assessment}
The localization AM $\alpha^*$ monitors the functional component $w^*$. This is represented as assessment node $\alpha$, functional node $w$ and functional trust $(\alpha, w) \in E_{ft}$ in the AG $S$ in Fig.~\ref{fig:assessment_graph}). Additionally the overall assessment node $A$ has trust in $\alpha$, represented by $(A, \alpha) \in E_{rt}$. The assessment is based on the works of Griebel et al. \cite{griebel_online_2023}. The changes in the position determined by the localization between each timestep are transformed into an opinion. Multiple consecutive opinions are fused together to form two sliding-window opinions, a short-term window (ST) and a long-term window (LT). The ST opinion is of fixed size. New opinions are fused into the ST opinion, while the oldest opinion is remove by unfusion and then fused into the LT opinion, where opinions decay over time. By comparing the ST and LT opinion sudden changes can be detected. By comparison to a reference opinion a degree of conflict can be calculated, which indicates if the relative localization over a specific time period follows the reference. In our case we facilitate odometry data as the reference opinion. The degree of conflict is then interpreted as opinion and used here as the assessment opinion $\omega_x^\beta$.

\begin{figure}[htbp]
    \centering
    \includegraphics[width=1.0\linewidth]{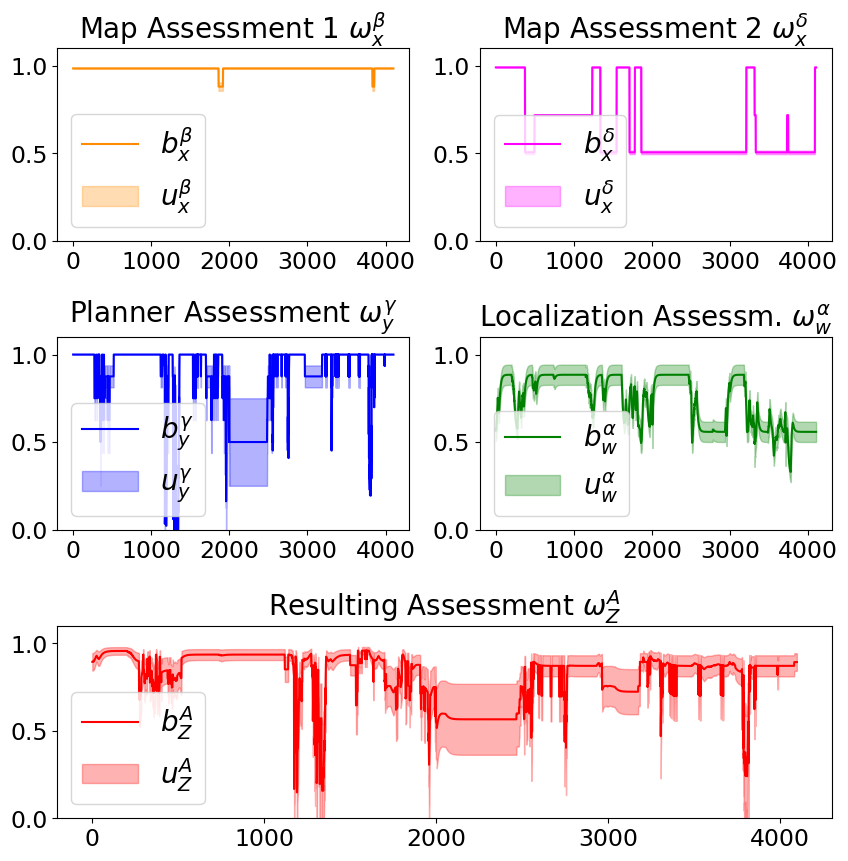}
    \caption{Belief and uncertainties of the assessment opinions for the map, planner and localization components over time. Belief and uncertainty of the resulting overall opinion $\omega_Z^A$ inferred with the AG $S$ in Fig.~\ref{fig:assessment_graph} depiced at the bottom, which clearly shows influence of the AMs.}
    \label{fig:result_plot}
\end{figure}
\subsection{Results}
Here, we briefly discuss results when applying an assessment with the presented framework on the AG $S$ in Fig.~\ref{fig:assessment_graph}. First we look at how trust discount, trust fusion and dependencies of input opinions affect resulting opinions. After that, we present the overall assessment with given assessment opinions.

\subsubsection{Trust Discount}
With a referral trust opinion $\omega_\gamma^A$ of $A$ on the AM $\gamma$ the influence of the assessment can be weighted (Sec.~\ref{sec:trusting_assessments}). In Fig.~\ref{fig:trust_discount_plot} a low trust (blue) at $t < 500$ is shown, that influences the opinion $\omega_y^\gamma$ (green) significantly resulting in a lower opinion $\omega_y^{[A;\gamma]}$ (red), while for $t \geq 500$ with high trust (blue) the resulting opinion (red) follows $\omega_y^\gamma$ (green) almost perfectly.

\addtolength{\textheight}{-1.15cm}
\subsubsection{Trust Fusion}
The fusion of trust is done when concurrent assessments have an opinion on the functionality of the same component (Sec.\ref{sec:concurrent_assessments}) or when incorporating the functional trust opinion of dependencies (Sec.~\ref{sec:opinions_on_dependencies}. In Fig.~\ref{fig:trust_fusion_plot} two concurrent assessment opinions $\omega_x^{[A;\beta]}, \omega_x^{[A;\delta]}$ (green, blue) are given and fused together into $\omega_x^{[A;\beta]\diamond [A;\delta]}$ (red). The resulting opinion reflects the belief of both opinions while its uncertainty decreases due to more evidence being present. The influence of input opinions is directed by their respective uncertainties.

\subsubsection{Dependencies}
The opinion on dependencies is calculated with deduction (Sec.~\ref{sec:opinions_on_dependencies}). In the case depicted in Fig.~\ref{fig:dependencies_plot} the set of opinions $\bm{\omega}_{\beta | \{u,w\}}$ that influence how the dependencies $u, w$ affect $\beta$ are set so, that only if both $u,w$ are operable $\beta$ is also operable with low uncertainty. In Fig.~\ref{fig:dependencies_plot} for $t < 500$ the dependent opinion $\omega_u^A$ on $u$ (green) shows operability with low uncertainty. The resulting opinion $\omega_{\beta || \{u,w\}}^A$ (red) is similar to the opinion $\omega_w^A$ on the other dependency $w$ (blue). For $t\geq500$, the lower belief and higher uncertainty of $\omega_u^A$ lead to a significant decrease in the belief and an increase in the uncertainty of the resulting opinion.

The overall assessment of the AG $S$ in Fig.~\ref{fig:assessment_graph} can be seen in Fig.~\ref{fig:result_plot}. The assessment opinions for the map, localization and planner components are depicted in the four plots at the top. The resulting overall opinion $\omega_Z^A$ (red) is shown in the bottom plot. It can be seen how all assessment opinions influence the overall assessment, whereas the planner assessment $\omega_y^\gamma$ (blue) has the most influence, since it is the component in the system from which all others are directly and indirectly are dependent.

\section{Outlook}
We presented a novel framework based on SL and SNs to infer a comprehensive functionality assessment of complex systems.  The transition from individual assessments to an overall judgment must address challenges such as concurrent evaluations, differential weighting of assessments, and the incorporation of system dependencies. The main advantage of our framework lies in its comprehensive bijective mapping between the system under test and our AG. By leveraging SL operators, we are able to flexibly infer an overall statement on the functionality of the system. We demonstrate the feasibility of our approach using real-world data from an AD vehicle software stack, and make our SL implementation publicly available in \cite{orf_subj_2025}.

\section*{Acknowledgment}
This work was developed within the framework of the Shuttle2X project, funded by the german Federal Ministry for Economic Affairs and Climate Action (BMWK) and the European Union, under the funding code 19S22001B.

\bibliographystyle{splncs04}
\bibliography{references.bib}

\begin{thebibliography}{10}
\providecommand{\url}[1]{\texttt{#1}}
\providecommand{\urlprefix}{URL }
\providecommand{\doi}[1]{https://doi.org/#1}

\bibitem{geissler_plausibility-based_2020}
Geissler, F., Unnervik, A., Paulitsch, M.: A {Plausibility}-{Based} {Fault} {Detection} {Method} for {High}-{Level} {Fusion} {Perception} {Systems}. IEEE Open Journal of Intelligent Transportation Systems  \textbf{1},  176--186 (2020)

\bibitem{griebel_online_2023}
Griebel, T., Heinzler, J., Buchholz, M., Dietmayer, K.: Online {Performance} {Assessment} of {Multi}-{Sensor} {Kalman} {Filters} {Based} on {Subjective} {Logic}. In: 2023 26th {International} {Conference} on {Information} {Fusion} ({FUSION}). IEEE, Charleston, SC, USA (Jun 2023)

\bibitem{heinrich_cocarnextgen_2024}
Heinrich, M., Zipfl, M., Uecker, M., Ochs, S., Gontscharow, M., Fleck, T., Doll, J., Schörner, P., Hubschneider, C., Zofka, M.R., Viehl, A., Zöllner, J.M.: Cocar nextgen: a multi-purpose platform for connected autonomous driving research (2024), \url{https://arxiv.org/abs/2404.17550}

\bibitem{hou_fault_2023}
Hou, W., Li, W., Li, P.: Fault {Diagnosis} of the {Autonomous} {Driving} {Perception} {System} {Based} on {Information} {Fusion}. Sensors  \textbf{23}(11) (May 2023)

\bibitem{fmea_2018}
{IEC 60812:2018}: Failure modes and effects analysis ({FMEA} and {FMECA}). Standard, International Electrotechnical Commission and others, Geneva, Switzerland (2018)

\bibitem{functional_safety_2018}
{ISO 26262:2018(en)}: Road vehicles — functional safety. Standard, International Organization for Standardization ({ISO}), Geneva, Switzerland (2018)

\bibitem{jin_hybrid_2024}
Jin, T., Zhang, C., Zhang, Y., Yang, M., Ding, W.: A {Hybrid} {Fault} {Diagnosis} {Method} for {Autonomous} {Driving} {Sensing} {Systems} {Based} on {Information} {Complexity}. Electronics  \textbf{13}(2) (Jan 2024)

\bibitem{josang_subjective_2016}
Jøsang, A.: Subjective {Logic}. Artificial {Intelligence}: {Foundations}, {Theory}, and {Algorithms}, Springer International Publishing, Cham (2016)

\bibitem{josang_subjective_2020}
Jøsang, A.: Subjective {Logic} (Oct 2020), \url{https://www.mn.uio.no/ifi/english/people/aca/josang/sl/}

\bibitem{lu_fault_2024}
Lu, Y., Li, G., Yue, Y., Wang, Z.: Fault {Detection} and {Data}-driven {Optimal} {Adaptive} {Fault}-tolerant {Control} for {Autonomous} {Driving} using {Learning}-based {SMPC}. IEEE Transactions on Intelligent Vehicles  (2024)

\bibitem{min_fault_2023}
Min, H., Fang, Y., Wu, X., Lei, X., Chen, S., Teixeira, R., Zhu, B., Zhao, X., Xu, Z.: A fault diagnosis framework for autonomous vehicles with sensor self-diagnosis. Expert Systems with Applications  \textbf{224} (Aug 2023)

\bibitem{orf_subj_2025}
Orf, S.: {SUBJ} - {Subjective} {Logic} {Library} (2025), \url{https://github.com/fzi-forschungszentrum-informatik/SUBJ/}

\bibitem{orf_modeling_2022}
Orf, S., Lambing, N., Ochs, S., Zofka, M.R., Zöllner, J.M.: Modeling {Localization} {Uncertainty} for {Enhanced} {Robustness} of {Automated} {Vehicles}. In: 2022 {IEEE} 18th {International} {Conference} on {Intelligent} {Computer} {Communication} and {Processing} ({ICCP}). pp. 175--182 (Sep 2022)

\bibitem{orf_modular_2024}
Orf, S., Ochs, S., Doll, J., Schotschneider, A., Heinrich, M., Zofka, M.R., Zöllner, J.M.: Modular {Fault} {Diagnosis} {Framework} for {Complex} {Autonomous} {Driving} {Systems}. In: 2024 {IEEE} 20th {International} {Conference} on {Intelligent} {Computer} {Communication} and {Processing} ({ICCP}). IEEE, Cluj-Napoca, Romania (Oct 2024)

\bibitem{poggenhans_lanelet2_2018}
Poggenhans, F., Pauls, J.H., Janosovits, J., Orf, S., Naumann, M., Kuhnt, F., Mayr, M.: Lanelet2: A high-definition map framework for the future of automated driving. In: 2018 21st International Conference on Intelligent Transportation Systems (ITSC). pp. 1672--1679 (2018)

\bibitem{seo_fail_2021}
Seo, K., Lee, J., Lee, J.y., Yi, K.: Fail {Safe} {Process} of {Vehicle} {Localization} for {Reliability} {Improvement} of {LV3} {Autonomous} {Driving}. International Journal of Automotive Technology  \textbf{22}(2),  529--535 (Apr 2021)

\bibitem{shen_integrated_2021}
Shen, Y., Xia, C., Jian, Z., Chen, S., Zheng, N.: An {Integrated} {Localization} {System} with {Fault} {Detection}, {Isolation} and {Recovery} for {Autonomous} {Vehicles}. In: 2021 {IEEE} {International} {Intelligent} {Transportation} {Systems} {Conference} ({ITSC}). pp. 84--91. IEEE, Indianapolis, IN, USA (Sep 2021)

\bibitem{stockem_novo_self-evaluation_2023}
Stockem~Novo, A., Hürten, C., Baumann, R., Sieberg, P.: Self-evaluation of automated vehicles based on physics, state-of-the-art motion prediction and user experience. Scientific Reports  \textbf{13}(1) (Aug 2023)

\bibitem{wang_risk_2022}
Wang, H., Lu, B., Li, J., Liu, T., Xing, Y., Lv, C., Cao, D., Li, J., Zhang, J., Hashemi, E.: Risk {Assessment} and {Mitigation} in {Local} {Path} {Planning} for {Autonomous} {Vehicles} {With} {LSTM} {Based} {Predictive} {Model}. IEEE Transactions on Automation Science and Engineering  \textbf{19}(4),  2738--2749 (Oct 2022)

\bibitem{yin_ieee_2020}
Yin, S., Kaynak, O., Reza~Karimi, H.: {IEEE} {Access} {Special} {Section} {Editorial}: {Data}-{Driven} {Monitoring}, {Fault} {Diagnosis} and {Control} of {Cyber}-{Physical} {Systems}. IEEE Access  \textbf{8} (2020)

\bibitem{youssef_general_2018}
Youssef, Y.M., Ota, D.: A general approach to health monitoring \& fault diagnosis of unmanned ground vehicles. In: 2018 {International} {Conference} on {Military} {Communications} and {Information} {Systems} ({ICMCIS}). IEEE, Warsaw, Poland (May 2018)

\end{thebibliography}
\end{document}